\documentclass[runningheads]{llncs}
\usepackage{graphicx}
\usepackage{amsmath}
\usepackage{amsfonts}
\usepackage{amssymb}
\usepackage{graphicx}%

\begin{document}
\title{A Generalized Stacking for Implementing Ensembles of Gradient Boosting Machines
\thanks{The reported study was funded by RFBR, project number 19-29-01004.}
}
\titlerunning{A Generalized Stacking Algorithm for Gradient Boosting Machines}

\author{Andrei V. Konstantinov \and Lev V. Utkin}
\authorrunning{A.V. Konstantinov and L.V. Utkin}
\institute{Peter the Great St.Petersburg Polytechnic University, Saint-Petersburg, Russia
\email{andrue.konst@gmail.com, lev.utkin@gmail.com}}
\maketitle              % typeset the header of the contribution
\begin{abstract}
The gradient boosting machine is one of the powerful tools for solving regression problems. In order to cope with its shortcomings, an approach for constructing ensembles of gradient boosting models is proposed. The main idea behind the approach is to use the stacking algorithm in order to learn a second-level meta-model which can be regarded as a model for implementing various ensembles of gradient boosting models. First, the linear regression of the gradient boosting models is considered as a simplest realization of the meta-model under condition that the linear model is differentiable with respect to its coefficients (weights). Then it is shown that the proposed approach can be simply extended on arbitrary differentiable combination models, for example, on neural networks which are differentiable and can implement arbitrary functions of gradient boosting models. Various numerical examples illustrate the proposed approach.

\end{abstract}

\section{Introduction}

One of the ways to enhance the machine learning models and to produce improved
results is to apply ensemble-based techniques which are based on combining a
set of the so-called base or weak models (classifiers, regressors)
\cite{Kuncheva-2004,Polikar-2012,Rokach-2019,Sagi-Rokach-2018,ZH-Zhou-2012}.
All approaches to combining models can be conditionally divided into three
main groups: bagging, stacking and boosting. The first group consists of
bagging methods \cite{Breiman-1996}, which are based on using bootstrapped
samples. One of the most well-known bagging models is the random forest
\cite{Breiman-2001} using a large number of randomly built classification or
regression decision trees whose predictions are combined to get the overall
random forest prediction. Random forests often use the combination of the
bagging and the random subspace method \cite{Ho-1998} for building trees. In
contrast to bagging methods, the boosting assigns weights to elements of a
training set in accordance with special rules. One of the first efficient
boosting algorithms is AdaBoost \cite{Freund-Shapire-97}. Among boosting
methods, we have to highlight the gradient boosting machines (GBMs)
\cite{Friedman-2001,Friedman-2002}. In accordance with these methods, the
training of each base model depends on models that have already been trained.
The interpretation of GBMs in terms of regression is the following. By using
the first guess as a prediction, the residuals are computed as differences
between guessed predictions and target variables. These residuals are used
instead of target variables to built the next base model, for example, a
regression tree which is used in turn to predict new residuals. The boosting
regression model is constructed by means of iterative computing the sum of all
previous regression trees and updating residuals to reflect changes in the
boosting regression model. In other words, a set of regression trees is
computed in the GBM such that each successive tree predicts the residuals of
the preceding trees given an arbitrary differentiable loss function
\cite{Sagi-Rokach-2018}. The gradient boosting has a lot of modifications. One
of the most popular modifications is the XGBoost \cite{Chen-Guestrin-2016}
which is much faster than other models. An efficient gradient boosting method
is the CatBoost \cite{Dorogush-etal-2018}.

Another interesting ensemble-based technique is stacking
\cite{Breiman-1996a,Wolpert-1992}. This technique is used to combine different
base models by means of a meta-learner that takes into account which base
model are reliable and which are not. One of the combination stacking models
is when outputs of the base models are used as training data for the
meta-learner to approximate the same target function. A detailed review of
stacking algorithms can be found in \cite{Sesmero-etal-15}.

It should be noted that the above division of the ensemble-based approaches is
rather rough because there are models which do not belong to these groups.
Moreover, there are a lot of models which can be viewed as a combination of
the above approaches, for example, the deep forest or gcForest which was
proposed by Zhou and Feng \cite{Zhou-Feng-2017a}. Due to many outperforming
properties of the deep forest and due to its architecture which is similar to
the multi-layer architecture of neural networks (NNs), several modifications
of the deep forest have been developed, for example,
\cite{Utkin-2019,Utkin-etal-2019d}. We have to point out also very interesting
combination of ideas of the gradient boosting and the deep forest, which is
called as the multi-layered gradient boosting decision tree forest
\cite{Feng-Yu-Zhou-2018}. It learns hierarchical distributed representations
by stacking several layers of regression gradient boosting decision trees as
its building blocks.

A lot of surveys have been published due to remarkable properties of
ensemble-based models
\cite{Fawagreh-etal-2014,Ferreira-Figueiredo-2012,Jurek-etal-2014,Kuncheva-2004,Polikar-2012,Ren-Zhang-Suganthan-2016,Rokach-2010,Wozniak-etal-2014,Yang-Yang-etal-2010}%
. Most ensemble-based models are thoroughly studied in Zhou's book
\cite{ZH-Zhou-2012}.

By returning to the GBMs, it should be pointed out that there are some
shortcomings of the technique, which are explicitly described by Natekin and
Knoll \cite{Natekin-Knoll-13}. One of the important shortcomings is that there
is currently no a fast and efficient model and the corresponding
implementation of the smooth continuous base learner that capture interactions
between variables which may play a crucial role in the particular predictive
model design. Moreover, the GBM can be regarded as a linear combination of
base models with some weights, and errors of the base models are correlated
for some examples of the training set such that the GBM overfits. In order to
overcome this shortcoming, we propose to extend the standard GBM towards
constructing ensembles of the models, i.e., ensembles of ensembles, to reduce
the impact of errors caused by overfitting. The ensemble is organized by using
the generalized stacking algorithm where inputs of the meta-model are
predictions of the ensemble of GBMs, and the meta-model can be implemented as
any differentiable machine learning model, i.e., the ensemble of ensembles of
GBMs is constructed by using arbitrary differentiable models, for example,
NNs. Moreover, the proposed approach allows us to reduce the number of tuning
parameters, that is many parameters become to be trainable. It is important to
note that the idea of combination of NNs and GBMs has been considered in
literature. In particular, Bilal \cite{Bilal-2019} proposed the so-called deep
gradient boosting where the GBM was incorporated into the NN backpropagation
algorithm at every layer of the NN for updating the NN weights. Another
combination of the GBM and the NN was proposed by Badirli at al.
\cite{Badirli-etal-2020} where the authors use shallow NNs as base learners in
the GBM. The same combination of the NN and XGBoost was proposed by
Weldegebriel et al. \cite{Weldegebriel-etal-2019}. Ideas of the NN and GBM
combination have been also studied by other authors, for example,
\cite{Bengio-etal-2006,Nitanda-Suzuki-2018}. However, our approach differs
from the available ones. We apply the NN as a possible tool for implementing
the stacking algorithm and the second-level ensemble of GBMs.

The peculiarities of the proposed approach open a door to develop a large
class of efficient ensembles of regression gradient boosting models. Various
numerical examples illustrate efficiency of the proposed ensembles of GBMs.

The paper is organized as follows. Sections 2 and 3 provide descriptions of
the standard regression problem statement and the GBM. An idea of using the
stacking algorithm for implementing the linear combination of GBMs is
considered in Section 4. An extension of the linear combination of GBMs
towards generalization of the proposed approach is given in Section 5.
Numerical examples with real data are provided in Section 6. Concluding
remarks can be found in Section 7.

\section{Regression problem statement}

Let us formally state the standard regression problem. Given $N$ training data
(examples, instances, patterns) $D=\{(x_{1},y_{1}),...,(x_{N},y_{N})\}$, in
which $x_{i}$ may belong to an arbitrary set $\mathcal{X}\subset\mathbb{R}%
^{m}$ and represents a feature vector involving $m$ features, and $y_{i}%
\in\mathbb{R}$ represents the observed output or target value such that
$y_{i}=f(x_{i})+\varepsilon$. Here $\varepsilon$ is the random noise with
expectation $0$ and unknown finite variance. Machine learning aims to
construct a regression model or an approximation $g$ of the function $f$ that
minimizes the expected risk or the expected loss function
\begin{equation}
L(f)=\mathbb{E}_{(x,y)\sim P}~l(y,g(x))=\int_{\mathcal{X}\times\mathbb{R}%
}l(y,g(x))\mathrm{d}P(x,y),\label{Imp_SVM16}%
\end{equation}
with respect to the function parameters. Here $P(x,y)$ is a joint probability
distribution of $x$ and $y$; the loss function $l(\cdot,\cdot)$ may be
represented, for example, as follows:
\begin{equation}
l(y,g(x))=\left(  y-g(x)\right)  ^{2}.
\end{equation}

There are many powerful machine learning methods for solving the regression
problem, including regression random forests
\cite{Biau-Scornet-2016,Breiman-2001}, the support vector regression
\cite{Smola-Scholkopf-2004}, etc. One of the powerful methods is the gradient
boosting \cite{Friedman-2002}, which will be considered below.

\section{The gradient boosting algorithm}

Let us consider the gradient boosting decision tree algorithm
\cite{Friedman-2002}. The algorithm is an iterative construction of a model as
an ensemble of base (weak) prediction models built in a stage-wise fashion
where each base model is constructed, based on data obtained using an ensemble
of models already built on previous iterations, as an approximation of the
loss function derivative. A model of the size $M$ is a linear combination of
$M$ base models:%

\begin{equation}
g_{M}(x)=\sum_{i=0}^{M}\gamma_{i}h_{i}(x),
\end{equation}
where $h_{i}$ is the $i$-th base model; $\gamma_{i}$ is the $i$-th coefficient
or the $i$-th base model weight.

The gradient boosting algorithm can be represented as the following steps:

\begin{enumerate}
\item Initialize the zero base model $h_{0}(x)$, for example, with the
constant value.

\item Calculate the residual $r_{i}^{(t)}$ as a partial derivative of the
expected loss function $L(x_{i},y_{i})$ at every points of the training set,
$i=1,...,N$.

\item Build the base model $h_{t}(x)$ as regression on residuals
$\{(x_{i},r_{i}^{(t)})\}$;

\item Find the optimal coefficient $\gamma_{t}$ at $h_{t}(x)$ regarding the
initial expected loss function (\ref{grad_bost_20});

\item Update the whole model $g_{t}(x)=g_{t-1}(x)+\gamma_{t}h_{t}(x)$;

\item If the stop condition is not fulfilled, go to step 2.
\end{enumerate}

Here the loss function depends on the machine learning problem solved
(classification or regression). Let us consider all the above in detail.
Suppose that $(M-1)$ steps produce the model $g_{M-1}(x)$. For constructing
the model $g_{M}(x)$, the model $h_{M}(x)$ has to be constructed, i.e., there
holds
\begin{equation}
g_{M}(x)=\sum_{t=1}^{M}\gamma_{t}h_{t}(x)=g_{M-1}(x)+\gamma_{M}h_{M}(x).
\end{equation}

The dataset for constructing the model $h_{M}(x)$ is chosen in such a way as
to approximate the expected loss function partial derivatives with respect to
the function of the already constructed model $g_{M-1}(x)$. Let us denote
residuals $r_{i}^{(M)}$ defined as the values of the loss function partial
derivative at point $g_{M-1}(x_{i})$ in the current iteration $M$,%
\begin{equation}
r_{i}^{(M)}=-\left.  \frac{\partial L(z,y_{i})}{\partial z}\right\vert
_{z=g_{M-1}(x_{i})}.
\end{equation}

By using the residuals, a new training set $D_{M}$ can be formed as follows:%
\begin{equation}
D_{M}=\left\{  \left(  x_{i},r_{i}^{(M)}\right)  \right\}  _{i=1}^{N},
\end{equation}
and the model $h_{M}$ can be constructed on $D_{M}$ by solving the following
optimization problem
\begin{equation}
\min\sum_{i=1}^{N}\left\Vert h_{M}(x_{i})-r_{i}^{(M)}\right\Vert
^{2}.\label{grad_bost_19}%
\end{equation}

Hence, an optimal coefficient $\gamma_{M}$ of the gradient descent can be
obtained as:%
\begin{equation}
\gamma_{M}=\arg\min_{\gamma}\sum_{i=1}^{N}L\left[  g_{M-1}(x)+\gamma
h_{M}(x_{i}),y_{i}\right]  .\label{grad_bost_20}%
\end{equation}

Then we get the following model at every point $x_{i}$ of the training set
\begin{equation}
g_{M}(x)=g_{M-1}(x)+\gamma_{M}h_{M}(x)\approx g_{M-1}(x)-\gamma_{M}\left.
\frac{\partial L(z,y_{i})}{\partial z}\right\vert _{z=g_{M-1}(x_{i})}.
\end{equation}

If the loss function is the squared difference, then minimizing of
(\ref{grad_bost_19}) corresponds to minimizing the loss function. This implies
that choice of the optimal step $\gamma_{M}$ is not required, and its value is
$1$. In order to reduce overfitting, the step $\gamma_{M}$ is reduced by its
multiplying by a constant called the learning rate. By introducing the
learning rate, it is possible to reduce the impact of model $h_{M}$ errors on
the ensemble error.

The above algorithm allows us to minimize the expected loss function by using
decision trees as base models. However, it requires to select the decision
tree parameters, for example, depths of trees, as well as the learning rate,
in order to simultaneously provide a high generalization and accuracy
depending on an specific task.

The gradient boosting algorithm is a powerful and efficient tool for solving
regression problems, which can cope with complex non-linear function
dependencies \cite{Natekin-Knoll-13}. However, it has a number of
shortcomings. One of them is caused by the \textquotedblleft
greedy\textquotedblright\ concept used in the model implementation. The GBM
$g_{M}(x)$ constructed using the above algorithm is itself an ensemble of base
models, i.e., a linear combination of base models with given weights
$\gamma_{i}$. However, each model is regarded as the \textquotedblleft
greedy\textquotedblright\ one, which means that errors of base models are
correlated in the worst case when the base models are poorly approximating
residuals for a set of points. As a result, the loss function can be minimized
in the vicinity of such points only by overfitting. Therefore, it makes sense
to construct ensembles of GBMs in order to reduce the variance of the error
caused by overfitting.

\section{An ensemble of GBMs}

The main idea behind the construction of a more accurate ensemble of GBMs is
to apply the stacking algorithm \cite{Wolpert-1992} which trains the
first-level learners using the original training dataset. The stacking
algorithm generates a new dataset for training the second-level learner
(meta-model) such that the outputs of the first-level learners are regarded as
input features for the second-level learner while the original labels are
still regarded as labels of new training data.

Let us compose an ensemble $E_{M}^{K}(x)$ of $K$ GBMs of the size $M$. It can
be represented as follows:%
\begin{equation}
E_{M}^{K}(x)=\left(  g_{M}^{(j)}(x)\right)  _{j=1}^{K}.
\end{equation}

First, we consider a linear regression model for implementing the meta-model
of the stacking algorithm, which is of the form:%
\begin{equation}
S_{w}(t)=t\cdot w+b=\sum_{j=1}^{K}t_{j}w_{j}+b,\label{grad_bost_28}%
\end{equation}
where $w=(w_{1},...,w_{K})\in\mathbb{R}^{K}$ is a vector of weights;
$t=(t_{1},...,t_{K})\in\mathbb{R}^{K}$ is a vector of the ensemble model
predictions; $b$ is the bias.

We will assume for simplicity purposes that the bias $b$ is zero. Weights of
the linear regression model can be computed by means of the standard
well-known approaches depending on the regularization method used (the $L_{1}
$ or $L_{2}$ norms). Suppose that a differentiable expected loss function
$L(\hat{y},y)$ is given, where $y$ is a true class label, $\hat{y}$ is the
model prediction. Let us set the initial approximation of the weights
$w^{(0)}$. We are searching for the optimal vector of weights using the
gradient descent:%
\begin{equation}
w^{(q)}=w^{(q-1)}-\alpha\frac{1}{N}\sum_{i=1}^{N}\left.  \nabla_{w}L\left(
S_{w}\left(  E_{M}^{K}(x_{i})\right)  ,y_{i}\right)  \right\vert
_{w=w^{(q-1)}}.
\end{equation}

Here $S_{w}\left(  E_{M}^{K}(x_{i})\right)  $ is the prediction of the
ensemble for the input feature vector $x_{i}$.

Note that the model $S_{w}(t)$ is differentiable both by the vector of weights
and by the vector $t$ of the ensemble predictions. It has been shown that the
use of the gradient boosting algorithm allows us to minimize the
differentiable loss functions by constructing new models approximating values
proportional to their derivatives. We will use this peculiarity to optimize
not only linear regression weights, but also the ensemble of GBMs.

Let us define a new differentiable loss function as follows:%
\begin{equation}
\mathcal{L}(t,y)=L\left(  S_{w}(t),y\right)  .\label{grad_bost_30}%
\end{equation}

Intuitively, such a loss functional \textquotedblleft hides\textquotedblright%
\ the linear regression block $S_{w}$. Since the gradient boosting algorithm
allows iteratively minimizing an arbitrary differentiable loss function, we
apply it to minimize $\mathcal{L}$. Each GBM in the ensemble minimizes the
corresponding loss function:%
\begin{equation}
\mathcal{L}_{i}(t_{i},y)=L\left(  S_{w}(t),y\right)  .\label{grad_bost_31}%
\end{equation}

By using gradient boosting for each fixed set of weights of the linear model,
we can construct such an ensemble of GBMs, which minimizes the loss function
(\ref{grad_bost_30}). Similarly, optimal weights of the linear model can be
determined for a fixed ensemble of GBMs. We combine these two steps into one
step, i.e., we simultaneously optimize weights of the linear model and
construct the ensemble of GBMs.

Suppose that an ensemble of $K$ GBMs\ is composed such that every its GBM is
initialized by a constant value, for example, by the mean value of the
corresponding target variable. Initial weights of the linear model are set as:%
\begin{equation}
w^{(0)}=\frac{1}{K}.
\end{equation}

Denote the residual of the $j$-th GBM for point $x_{i}$ at step $q$ as
$r_{i,j}^{q}$, and differentiate the loss function $\mathcal{L}$ as:%
\begin{equation}
r_{i,j}^{q}=-\left.  \frac{\partial\mathcal{L}(t,y_{i})}{\partial t_{j}%
}\right\vert _{t=E_{q-1}^{K}(x_{i})}=-\left.  \frac{\partial L(z,y_{i}%
)}{\partial z}\right\vert _{z=S_{w}(t)}\cdot\left.  \frac{\partial S_{w}%
(t)}{\partial t_{j}}\right\vert _{t=E_{q-1}^{K}(x_{i})}.
\end{equation}

The partial derivative with respect to the $j$-th component of the linear
regression model is nothing else but the weight $w_{j}$ in (\ref{grad_bost_28}%
) corresponding to this component. Hence, there holds:%
\begin{equation}
r_{i,j}^{q}=-\left.  \frac{\partial L(z,y_{i})}{\partial z}\right\vert
_{z=S_{w}(t)}\cdot w_{j}^{(q-1)}.\label{grad_bost_33}%
\end{equation}

Let us construct a function which approximates residuals as:
\begin{equation}
h_{q}^{j}(x)=\arg\min_{h\in\mathcal{F}}\sum_{i=1}^{N}\left\Vert h(x_{i}%
)-r_{i,j}^{q}\right\Vert ^{2},\label{grad_bost_34}%
\end{equation}
where $\mathcal{F}$ is a set of admissible functions.

We simultaneously optimize weights of the linear model and each GBM in the
ensemble as follows:
\begin{equation}
\left\{
\begin{array}
[c]{c}%
w^{(q)}=w^{(q-1)}-\alpha\dfrac{1}{N}%
%TCIMACRO{\dsum \limits_{i=0}^{N}}%
%BeginExpansion
{\displaystyle\sum\limits_{i=0}^{N}}
%EndExpansion
\left.  \nabla_{w}L\left(  S_{w}\left(  E_{M}^{K}(x_{i})\right)
,y_{i}\right)  \right\vert _{w=w^{(q-1)}},\\
g_{q}^{(j)}=g_{q-1}^{(j)}+\gamma_{q}^{(j)}h_{q}^{j}(x).
\end{array}
\right.
\end{equation}

For simplicity, $\gamma_{q}^{(j)}$ will be taken identical for all models and
iterations. In sum, the learning algorithm consists of the following steps:

\begin{enumerate}
\item Initialize an ensemble of $K$ GBMs.

\item Initialize model weights $S_{w}$.

\item Until the stop condition is fulfilled, at step $q$:

\begin{enumerate}
\item calculate the partial derivative of the loss function by weights of the
linear model;

\item calculate the residuals $r_{i,j}^{q}$;

\item construct base models $h_{q}^{j}(x)$ approximating the corresponding
residues $r_{i,j}^{q}$.
\end{enumerate}
\end{enumerate}

The resulting model will be called as an adaptive ensemble of GBMs because
each member of the GBM ensemble adapts to a new loss function corresponding to
each iteration.

It should be noted that results of the traditional gradient boosting algorithm
are strongly influenced by the choice of parameters of base models (decision
trees). For every specific problem, the most appropriate parameters exist.
However, it is necessary to construct the GBM many times to find the
parameters and to apply one of the evaluation methods, for example, the
cross-validation method. In the adaptive ensemble of GBMs, many different
parameters of the base models can be immediately included as follows. In each
GBM of the ensemble, we use a unique set of parameters corresponding to the
GBM throughout the entire training of the ensemble. As a result of training,
the largest weights are assigned to models with the most appropriate parameters.

As a rule, decision trees in traditional gradient boosting algorithms are used
as base models. In order to construct a model $h_{q}^{j}(x)$ approximating the
residuals $r_{i,j}^{q}$, a tree is built that minimizes the quadratic norm of
the difference between the model predictions and the residual
(\ref{grad_bost_34}). The decision tree model approximating residuals
implements a piecewise constant function with a small number of unique values.
That is, at each iteration of the ensemble training, accuracy of a prediction
of the partial derivative of the loss function (\ref{grad_bost_33}) is not
high. Moreover, it is difficult to control the accuracy of the approximation
under condition of a fixed tree depth for each specific GBM. Hence, ensembles
of GBMs with a larger depth of trees will learn \textquotedblleft
faster\textquotedblright\ than GBM with a smaller depth, although trees of a
smaller depth may be more preferable for a specific task. The basic idea
behind the problem solution is to use more complex base models, namely, GBMs.
Such models allow us to approximate the residuals with a higher accuracy by
using trees of a fixed depth. It is important to note that the ensemble of
GBMs, which uses GBMs as base models, is also a linear combination of decision
trees, however, the algorithm at each step more accurately approximates the
partial derivative of the loss function (\ref{grad_bost_31}).

Let us consider various ways for initializing an ensemble of deep GBMs:

\begin{enumerate}
\item An exact initialization using the training set: each GBM is constructed
by optimizing the loss function $L$ directly over a certain number of
iterations $M_{init}$. The larger the number of iterations, the greater the
correlation between the model predictions.

\item A random initialization: each GBM includes only a single base model
built on the basis of the training set where values {}{}of free variables
repeat those of $D$, and reference values {}{}of the target variable are
selected as observations of a random variable from the normal distribution
whose parameters, the mean and the variance, coincide with the sample mean and
the sample variance of the reference target variable.

\item An exact initialization using subsets of the training set: in order to
get the most diverse GBMs in the ensemble (the diversity reduces the
correlation of the model prediction errors), the GBMs can be constructed on
pairwise disjoint subsets of the training set.

\item An initialization with the average value of the target variable.
\end{enumerate}

Weights of the linear model can be initialized with optimal values under
condition that the ensemble of GBMs is fixed.

\section{Generalization of the GBM ensemble}

Note that only a few conditions have to be fulfilled for training the
aggregate model by means of the proposed algorithm:

\begin{itemize}
\item the loss function $L$ has to be differentiable;

\item the model $S_{w}(t)$ has to be differentiable with respect to the
parameter $w$ and the argument $t$.
\end{itemize}

This implies that the obtained approach can be generalized by replacing the
linear regression model with an arbitrary differentiable model $\mathcal{S}%
_{\theta}^{K}(t)$, whose input is a vector of size $K$. The learning procedure
for the model $\mathcal{S}_{\theta}^{K}(t)$ can be carried out not only by
gradient descent, but also by another algorithm, for example, by means of the
support vector method. However, the model in this case has to be rebuilt after
each update of the ensemble of GBMs. Therefore, we will consider only
differentiable models $\mathcal{S}_{\theta}^{K}(t)$ from the class of NNs of
forward propagation $\mathcal{F}$:%
\begin{equation}
\mathcal{F}=\left\{  f:\mathbb{R}^{I}\times\Theta\rightarrow\mathbb{R}%
^{T}\right\}  ,
\end{equation}
where $I$ is the dimensionality of the NN input space; $\Theta$ is the set of
admissible parameters of the NN; $T$ is the dimensionality of the target variable.

Any method of initializing the GBM ensemble among the considered above can be
used for the case $T=1$, and the dimension of the input space is equal to the
number of models: $I=K$. If $T>1$, then it is possible to exactly initialize
every GBM in the ensemble using the training set only by constructing models
with several outputs such that every GBM is a function whose values {}{}are in
a space of dimension $T$. In this case, there holds $I=T\cdot K$, and the
prediction of the GBM ensemble is the concatenation of predictions of every
GBM. In the case of large values {}{}of $T$, construction of the GBM ensemble
with several outputs may be computationally expensive. For example, if the
target variable corresponds to images of sizes $100\times100$, and the
ensemble contains $100$ GBMs, then the dimension of the input space of the NN
will be $10^{6}$. When the size of the training set is equal to $1000$ in this
case, the number of values {}{}that should be fed to the NN input is one
billion. Therefore, it makes sense in such cases to use the random
initialization which allows us to directly control the value of $I$ by setting
the number of models $K$.

We can construct a model on the basis of the proposed algorithm using both the
advantages of learning algorithms for base models, for example, for decision
trees, and the advantages of differentiable models, including the NN. The GBM
ensemble is actually a mapping from the original feature space into a new
intermediate space of dimension $I$, which is more informative in the context
of a specific problem. Moreover, such the GBM ensemble has the following
advantages of traditional ensembles based on decision trees:

\begin{itemize}
\item possibility of training the ensemble of trees on a sample of small dimension;

\item interpretability of models, in particular, an assessment of the effects
of specific features of input data on predictions;

\item availability of deterministic algorithms for constructing decision trees;

\item lack of assumptions about the existence of a linear relationship between features.
\end{itemize}

Note that the aggregate model inherits a part of the above properties, namely
the interpretability and the lack of assumptions about the existence of a
linear relationship between features, while the useful properties of the NN
are potentially preserved, including generating high-dimensional outputs (for
example, images), simultaneous solving several types of tasks (multi-task
learning), parameterizing the loss function of a neural network, and so on.
But it is very important that the number of NN parameters can be reduced, and,
as a result, the NN can be trained on small samples because processing of the
initial features is carried out to the NN.

\section{Numerical experiments}

In order to illustrate the proposed approach, we investigate the model for
real data sets from the R Package \textquotedblleft
datamicroarray\textquotedblright\ which contains DNA microarrays. Table
\ref{t:DNA_datasets} is a brief introduction about the investigated datasets,
while more detailed information can be found from, respectively, the data
resources. Table \ref{t:DNA_datasets} shows the number of features $m$ for the
corresponding data set, the number of examples $n$ and the number of classes
$C$. All these datasets are for solving the classification task. It can be
seen from Table \ref{t:DNA_datasets} that the number of features of every
dataset is much more than the number of training examples. Accuracy measure
$A$ used in numerical experiments is the proportion of correctly classified
cases on a sample of data. To evaluate the average accuracy, we perform a
cross-validation with $100$ repetitions, where in each run, we randomly select
$n_{\text{tr}}=3n/4$ training data and $n_{\text{test}}=n/4$ testing data. %

\begin{table}[tbp] \centering
\caption{A brief introduction about the DNA microarray data sets}%
\begin{tabular}
[c]{ccccc}\hline
Data set & Type & $m$ & $n$ & $C$\\\hline
Alon & Colon Cancer & $2000$ & $62$ & $2$\\\hline
Borovecki & Huntington's Disease & $22283$ & $31$ & $2$\\\hline
Chin & Breast Cancer & $22215$ & $118$ & $2$\\\hline
Chowdary & Breast Cancer & $22283$ & $104$ & $2$\\\hline
Golub & Leukemia & $7129$ & $72$ & $3$\\\hline
Gravier & Breast Cancer & $2905$ & $168$ & $2$\\\hline
Pomeroy & CNS Disorders & $7128$ & $60$ & $2$\\\hline
Nakayama & Sarcoma & $22283$ & $105$ & $10$\\\hline
Sorile & Breast Cancer & $456$ & $85$ & $5$\\\hline
Singh & Prostate Cancer & $12600$ & $102$ & $2$\\\hline
\end{tabular}
\label{t:DNA_datasets}%
\end{table}%

Numerical results of the DNA microarray classification are shown in Table
\ref{t:DNA_results}. Four models are compared: the linear ensemble of GBMs
(Linear GBM) in accordance with (\ref{grad_bost_28}); combination of the
NN\ and ensemble of GBMs (NN+GBM), the standard GBM and the random forest with
the numbers of decision trees equal to $100$ or $1000$. The learning rates
$0.1$ or $0.01$ are taken for training GBMs. Ensembles of GBMs consist of $20$
machines with depths of trees from $2$ to $21$. The epoch number is $10$, and
the learning rate of every ensemble is $0.05$. The fully connected NN having
$3$ layers of size $10$ with tanh as an activation functions. The best
performance for each dataset is shown in bold. It can be seen from Table
\ref{t:DNA_results} that the combinations of the NN and ensemble of GBMs
provide better results for most datasets. Moreover, $7$ datasets from $10$
ones show outperforming results by using the proposed approach ($2$ datasets
by using Linear GBM and $5$ datasets by using NN+GBM).%

\begin{table}[tbp] \centering
\caption{Comparison of four models on the DNA microarray datasets}
\begin{tabular}
[c]{ccccc}\hline
Data set & Linear GBM & NN + GBM & GBM & Random Forest\\\hline
Alon & $0.708$ & $\mathbf{0.833}$ & $0.771$ & $0.762$\\\hline
Borovecki & $0.979$ & $\mathbf{1.000}$ & $0.958$ & $0.925$\\\hline
Chin & $\mathbf{0.978}$ & $0.894$ & $0.900$ & $0.933$\\\hline
Chowdary & $0.987$ & $\mathbf{1.000}$ & $0.962$ & $0.992$\\\hline
Golub & $0.861$ & $\mathbf{0.991}$ & $0.889$ & $0.911$\\\hline
Gravier & $0.730$ & $0.766$ & $\mathbf{0.778}$ & $0.748$\\\hline
Pomeroy & $0.511$ & $0.422$ & $\mathbf{0.556}$ & $0.453$\\\hline
Nakayama & $0.577$ & $\mathbf{0.590}$ & $0.526$ & $0.585$\\\hline
Sorile & $0.627$ & $0.714$ & $0.683$ & $\mathbf{0.857}$\\\hline
Singh & $\mathbf{0.929}$ & $0.897$ & $0.897$ & $0.862$\\\hline
\end{tabular}
\label{t:DNA_results}%
\end{table}%

In order to study the proposed approach for solving regression problems, we
apply datasets described in Table \ref{t:regres_datasets}. The datasets are
taken from open sources, in particular, Boston Diabetes, Longley can be found
in the corresponding R Packages; HouseART can be found in the Kaggle platform;
Friedman 1 and 2 are described at site:
https://www.stat.berkeley.edu/\symbol{126}breiman/bagging.pdf;  Regression is
available in package \textquotedblleft Scikit-Learn\textquotedblright. %

\begin{table}[tbp] \centering
\caption{A brief introduction about the regression data sets}%
\begin{tabular}
[c]{cccc}\hline
Data set & Abbreviation & $m$ & $n$\\\hline
House Prices: Advanced Regression Techniques & HouseART & $79$ &
$1460$\\\hline
ML housing dataset & Boston & $13$ & $506$\\\hline
Diabetes & Diabetes & $10$ & $442$\\\hline
Longley's Economic Regression Data & Longley & $7$ & $16$\\\hline
Friedman 1 & Friedman 1 & $10$ & $100$\\\hline
Friedman 2 & Friedman 2 & $4$ & $100$\\\hline
Scikit-Learn Regression & Regression & $100$ & $100$\\\hline
\end{tabular}
\label{t:regres_datasets}%
\end{table}

Numerical results in the form of the mean squared errors for the regression
datasets are shown in Table \ref{t:regression_results}. We again use four
models described above. Ensembles of GBMs consist of $100$ machines now. The
NN having $3$ layers of size $20$ are used. Other parameters of numerical
experiments are the same as in the previous experiments. It can be seen from
Table \ref{t:regression_results} that the combinations of the NN and ensemble
of GBMs provide better results for $4$ datasets from $7$ ones. 

We can conclude after analyzing the numerical results that the proposed
approach provides outperforming results for cases of small datasets (see, for
example, most DNA microarray datasets and the Longley dataset). This implies
that ensembles of GBMs partially solve the problem of overfitting, which takes
place for datasets with the small number of training examples. %

\begin{table}[tbp] \centering
\caption{Comparison of four models on the  regression datasets}%
\begin{tabular}
[c]{ccccc}\hline
Data set & Linear GBM & NN + GBM & GBM & Random Forest\\\hline
HouseART & $\mathbf{6.43\times10^{8}}$ & $2.02\times10^{8}$ & $8.02\times
10^{8}$ & $9.26\times10^{8}$\\\hline
Boston & $1.42\times10^{1}$ & $\mathbf{1.16\times10^{1}}$ & $1.47\times10^{1}$
& $1.64\times10^{1}$\\\hline
Diabetes & $4.06\times10^{3}$ & $3.73\times10^{3}$ & $\mathbf{3.48}%
$\textbf{$\times10^{3}$} & $3.74\times10^{3}$\\\hline
Longley & $\mathbf{1.01\times10^{0}}$ & $6.04\times10^{0}$ & $1.91\times
10^{0}$ & $1.65\times10^{0}$\\\hline
Friedman 1 & $8.07\times10^{0}$ & $1.13\times10^{1}$ & $\mathbf{6.82\times
10^{0}}$ & $9.41\times10^{0}$\\\hline
Friedman 2 & $\mathbf{2.06}$\textbf{$\times10^{3}$} & $6.35\times10^{3}$ &
$2.40\times10^{3}$ & $3.85\times10^{3}$\\\hline
Regression & $1.15\times10^{4}$ & $1.84\times10^{4}$ & $\mathbf{1.06}%
$\textbf{$\times10^{4}$} & $1.34\times10^{4}$\\\hline
\end{tabular}
\label{t:regression_results}%
\end{table}%
\qquad\qquad\qquad

\section{Conclusion}

A new approach for combining GBMs by using the stacking algorithm has been
proposed in the paper. It has many advantages in comparison with the GBM as
well as with deep differentiable models such as NNs:

\begin{itemize}
\item the \textquotedblleft greedy\textquotedblright\ stacking algorithm of
GBMs does not guarantee an achievement of the loss function optimum because
the optimization procedure is carried out in turn. The simultaneous
optimization solves this problem;

\item NNs consider linear combinations of input features, which lead to a
serious problem of overfitting when working with tabular data consisting of
features of different nature, for example, mass and length, as well as by a
large number of features and small sizes of training samples. The proposed
approach allows us to process features using decision trees and to construct
arbitrarily deep models taking advantages of NNs.
\end{itemize}

It should be noted that many important questions and studies remain outside
the scope of our study in this paper. In particular, it is interesting to
consider various types of regularization which could improve the models.
Moreover, it is interesting to consider a procedure which removes a training
example from the gradient descent procedure when a current residual
corresponding to the example is smaller than some threshold. This improvement
may reduce the learning time and increase the model accuracy. The above
questions can be regarded as directions for further research.

\bibliographystyle{splncs04}
\bibliography{Boosting,Classif_bib,Deep_Forest,MYBIB,MYUSE}

\begin{thebibliography}{10}
\providecommand{\url}[1]{\texttt{#1}}
\providecommand{\urlprefix}{URL }
\providecommand{\doi}[1]{https://doi.org/#1}

\bibitem{Badirli-etal-2020}
Badirli, S., Liu, X., Xing, Z., Bhowmik, A., Keerthi, S.: Gradient boosting
  neural networks: Grownet (Feb 2020), arXiv:2002.07971

\bibitem{Bengio-etal-2006}
Bengio, Y., Roux, N., Vincent, P., Delalleau, O., Marcotte, P.: Convex neural
  networks. In: Advances in neural information processing systems. pp. 123--130
  (2006)

\bibitem{Biau-Scornet-2016}
Biau, G., Scornet, E.: A random forest guided tour. Test  \textbf{25}(2),
  197--227 (2016)

\bibitem{Bilal-2019}
Bilal, E.: Deep gradient boosting -- layer-wise input normalization of neural
  networks (Oct 2019), arXiv:1907.12608

\bibitem{Breiman-1996}
Breiman, L.: Bagging predictors. Machine Learning  \textbf{24}(2),  123--140
  (1996)

\bibitem{Breiman-1996a}
Breiman, L.: Stacked regressions. Machine Learning  \textbf{24}(1),  49--64
  (1996)

\bibitem{Breiman-2001}
Breiman, L.: Random forests. Machine learning  \textbf{45}(1),  5--32 (2001)

\bibitem{Chen-Guestrin-2016}
Chen, T., Guestrin, C.: Xgboost: A scalable tree boosting system. In:
  Proceedings of the 22nd {ACM SIGKDD} International Conference on Knowledge
  Discovery and Data Mining. pp. 785--794. ACM, New York, NY (2016)

\bibitem{Dorogush-etal-2018}
Dorogush, A., Ershov, V., Gulin, A.: Catboost: gradient boosting with
  categorical features support (Oct 2018), arXiv:1810.11363

\bibitem{Fawagreh-etal-2014}
Fawagreh, K., Gaber, M., Elyan, E.: Random forests: from early developments to
  recent advancements. Systems Science \& Control Engineering  \textbf{2}(1),
  602--609 (2014)

\bibitem{Feng-Yu-Zhou-2018}
Feng, J., Yu, Y., Zhou, Z.H.: Multi-layered gradient boosting decision trees.
  In: Advances in Neural Information Processing Systems. pp. 3551--3561. Curran
  Associates, Inc. (2018)

\bibitem{Ferreira-Figueiredo-2012}
Ferreira, A., Figueiredo, M.: Boosting algorithms: A review of methods, theory,
  and applications. In: Zhang, C., Ma, Y. (eds.) Ensemble Machine Learning:
  Methods and Applications, pp. 35--85. Springer, New York (2012)

\bibitem{Freund-Shapire-97}
Freund, Y., Schapire, R.: A decision theoretic generalization of on-line
  learning and an application to boosting. Journal of Computer and System
  Sciences  \textbf{55}(1),  119--139 (1997)

\bibitem{Friedman-2001}
Friedman, J.: Greedy function approximation: A gradient boosting machine.
  Annals of Statistics  \textbf{29},  1189--1232 (2001)

\bibitem{Friedman-2002}
Friedman, J.: Stochastic gradient boosting. Computational statistics \& data
  analysis  \textbf{38}(4),  367--378 (2002)

\bibitem{Ho-1998}
Ho, T.: The random subspace method for constructing decision forests. IEEE
  Transactions on Pattern Analysis and Machine Intelligence  \textbf{20}(8),
  832--844 (1998)

\bibitem{Jurek-etal-2014}
Jurek, A., Bi, Y., Wu, S., Nugent, C.: A survey of commonly used ensemble-based
  classification techniques. The Knowledge Engineering Review  \textbf{29}(5),
  551--581 (2014)

\bibitem{Kuncheva-2004}
Kuncheva, L.: Combining Pattern Classifiers: Methods and Algorithms.
  Wiley-Interscience, New Jersey (2004)

\bibitem{Natekin-Knoll-13}
Natekin, A., Knoll, A.: Gradient boosting machines, a tutorial. Frontiers in
  neurorobotics  \textbf{7}(Article 21),  1--21 (2013)

\bibitem{Nitanda-Suzuki-2018}
Nitanda, A., Suzuki, T.: Functional gradient boosting based on residual network
  perception (Feb 2018), arXiv:1802.09031

\bibitem{Polikar-2012}
Polikar, R.: Ensemble learning. In: Zhang, C., Ma, Y. (eds.) Ensemble Machine
  Learning: Methods and Applications, pp. 1--34. Springer, New York (2012)

\bibitem{Ren-Zhang-Suganthan-2016}
Ren, Y., Zhang, L., Suganthan, P.N.: Ensemble classification and
  regression-recent developments, applications and future directions [review
  article]. IEEE Computational Intelligence Magazine  \textbf{11}(1),  41--53
  (2016)

\bibitem{Rokach-2010}
Rokach, L.: Ensemble-based classifiers. Artificial Intelligence Review
  \textbf{33}(1-2),  1--39 (2010)

\bibitem{Rokach-2019}
Rokach, L.: Ensemble Learning: Pattern Classification Using Ensemble Methods,
  vol.~85. World Scientific (2019)

\bibitem{Sagi-Rokach-2018}
Sagi, O., Rokach, L.: Ensemble learning: A survey. WIREs Data Mining and
  Knowledge Discovery  \textbf{8}(e1249),  1--18 (2018)

\bibitem{Sesmero-etal-15}
Sesmero, M., Ledezma, A., Sanchis, A.: Generating ensembles of heterogeneous
  classifiers using stacked generalization. WIREs Data Mining and Knowledge
  Discovery  \textbf{5},  21--34 (2015)

\bibitem{Smola-Scholkopf-2004}
Smola, A., Scholkopf, B.: A tutorial on support vector regression. Statistics
  and Computing  \textbf{14},  199--222 (2004)

\bibitem{Utkin-2019}
Utkin, L.: An imprecise extension of svm-based machine learning models.
  Neurocomputing  \textbf{331},  18--32 (2019).
  \doi{10.1016/j.neucom.2018.11.053}

\bibitem{Utkin-etal-2019d}
Utkin, L., Konstantinov, A., Meldo, A., Ryabinin, M., Chukanov, V.: A deep
  forest improvement by using weighted schemes. In: Proceedings of the 24th
  Conference of Open Innovations Association FRUCT. pp. 451--456. IEEE, Moscow,
  Russia (2019). \doi{10.23919/FRUCT.2019.8711886}

\bibitem{Weldegebriel-etal-2019}
Weldegebriel, H., Liu, H., Haq, A., Bugingo, E., Zhang, D.: A new hybrid
  convolutional neural network and extreme gradient boosting classifier for
  recognizing handwritten ethiopian characters. IEEE Access  \textbf{8},
  17804--17818. (2019)

\bibitem{Wolpert-1992}
Wolpert, D.: Stacked generalization. Neural networks  \textbf{5}(2),  241--259
  (1992)

\bibitem{Wozniak-etal-2014}
Wozniak, M., Grana, M., Corchado, E.: A survey of multiple classifier systems
  as hybrid systems. Information Fusion pp. 3--17 (2014)

\bibitem{Yang-Yang-etal-2010}
Yang, P., Yang, E., Zhou, B., Zomaya, A.: A review of ensemble methods in
  bioinformatics. Current Bioinformatics  \textbf{5}(4),  296--308 (2010)

\bibitem{ZH-Zhou-2012}
Zhou, Z.H.: Ensemble Methods: Foundations and Algorithms. CRC Press, Boca Raton
  (2012)

\bibitem{Zhou-Feng-2017a}
Zhou, Z.H., Feng, J.: Deep forest: Towards an alternative to deep neural
  networks. In: Proceedings of the 26th International Joint Conference on
  Artificial Intelligence (IJCAI'17). pp. 3553--3559. AAAI Press, Melbourne,
  Australia (2017)

\end{thebibliography}

\end{document}